\documentclass[conference]{IEEEtran}
\IEEEoverridecommandlockouts
\usepackage{cite}
\usepackage{amsmath,amssymb,amsfonts}
\usepackage{algorithmic}
\usepackage{graphicx}
\usepackage{textcomp}
\usepackage{xcolor}
\usepackage{subcaption}
\usepackage{flushend}

\def\BibTeX{{\rm B\kern-.05em{\sc i\kern-.025em b}\kern-.08em
    T\kern-.1667em\lower.7ex\hbox{E}\kern-.125emX}}
\begin{document}

\title{Efficient Model Compression for Hierarchical Federated Learning\\

}

\author{Xi~Zhu$^{1}$,~Songcan~Yu$^{2}$,~Junbo~Wang$^{2}$\textsuperscript{*},\thanks{\textsuperscript{*} is the corresponding author.} \thanks{This article has been accepted by IEEE ICC 2023.}
~Qinglin~Yang$^{2}$\\
 \normalsize $^1$School of Systems Science and Engineering, Sun Yat-Sen University, Guangzhou, China\\$^2$School of Intelligent Systems Engineering, Sun Yat-Sen University, Guangzhou, China\\
\{zhux66, yusc\}@mail2.sysu.edu.cn, \{wangjb33,yangqlin6\}@mail.sysu.edu.cn}

\maketitle

\begin{abstract}
Federated learning (FL), as an emerging collaborative learning paradigm, has garnered significant attention due to its capacity to preserve privacy within distributed learning systems. In these systems, clients collaboratively train a unified neural network model using their local datasets and share model parameters rather than raw data, enhancing privacy. Predominantly, FL systems are designed for mobile and edge computing environments where training typically occurs over wireless networks. Consequently, as model sizes increase, the conventional FL frameworks increasingly consume substantial communication resources. To address this challenge and improve communication efficiency, this paper introduces a novel hierarchical FL framework that integrates the benefits of clustered FL and model compression. We present an adaptive clustering algorithm that identifies a core client and dynamically organizes clients into clusters. Furthermore, to enhance transmission efficiency, each core client implements a local aggregation with compression (LC aggregation) algorithm after collecting compressed models from other clients within the same cluster. Simulation results affirm that our proposed algorithms not only maintain comparable predictive accuracy but also significantly reduce energy consumption relative to existing FL mechanisms.
\end{abstract}


\section{Introduction}

\par With the breakthrough of artificial intelligence, deep learning \cite{lecun2015deep} has been successfully applied in the field of IoT systems, such as smart vehicle  \cite{du2020federated, fatemi2023four}, and smart industrial \cite{rani2024secure}. However, the conventional centralized learning scheme adopts the way of uploading those data to the remote cloud for processing, which encounter many issues, including privacy leakage, high transmission cost, and so on. 

\begin{figure}[htb]
\centering
\includegraphics[width=2.4in]{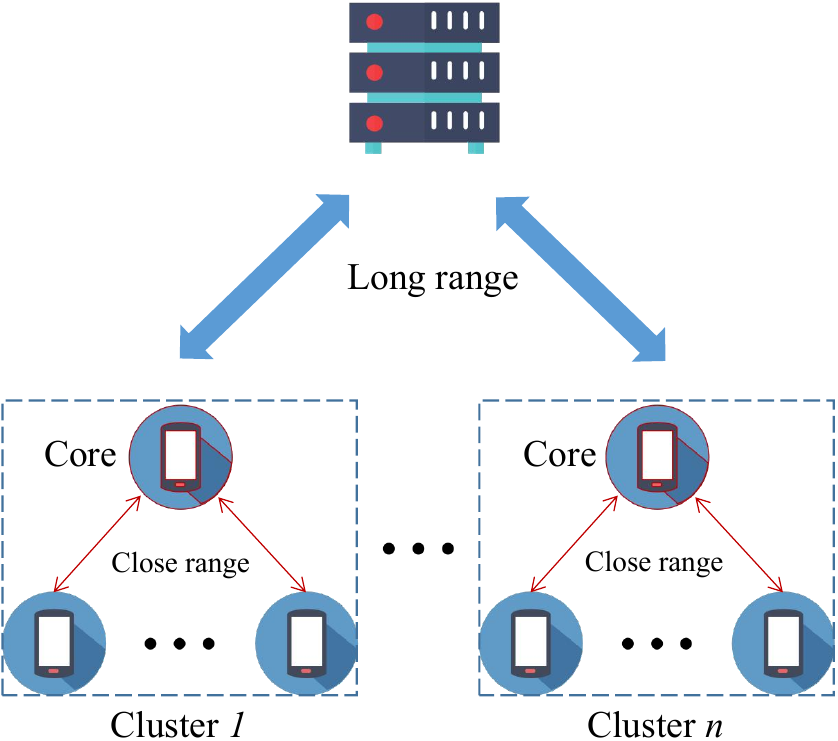}
\caption{Model compression based hierarchical FL}
\label{system_model}
\end{figure}

\par Fortunately, to train models without intruding on privacy information and alleviate the network bandwidth burden, federated learning (FL) has been proposed \cite{yang2019federated} as an efficient collaboratively solution. Reviewing the previous works of FL, the dominant framework is mainly based on the parameter server. It includes two layers i.e., the server layer and client layer in which clients utilize their local dataset to collaboratively train a prediction model, thereby keeping all the local training data in private.

\par Although FL reduces the communication cost and protects the privacy of the local data to some extent. This parameter server based two-layer framework \cite{wei2021lightweight} still incurs many difficulties. For instance, if the client and server are far apart, communication will be inefficient. Even if only the model is transmitted in the FL framework, the size of the model is still too large to meet the requirement of green communication since the AlexNet model and VGG-16 model are 240 MB and 552 MB \cite{han2015deep}, respectively. 

\par To reduce communication costs, model compression techniques like model pruning \cite{wu2022improved}, quantification \cite{du2020high}, and other methods such as knowledge distillation (KD) \cite{hinton2015distilling} and tensor decomposition \cite{yin2021towards} are used to decrease the model size transmitted. While these approaches can slightly reduce network traffic, they still result in high computation latency for aggregation due to not considering the number of aggregating clients.  


\par Given the outlined challenges, there is an urgent need for more robust solutions. Inspired by cloud-fog computing principles, which facilitate cooperative task processing among clients \cite{chen2018fog}, we propose a novel hierarchical FL framework that employs model compression to significantly cut communication costs between clients and servers.

\par This new framework not only reduces the number of models transmitted to the server but also cuts computation latency by integrating model pruning and perturbation techniques. Additionally, clients are adaptively grouped into clusters, with a core client in each cluster tasked with aggregating all local models beforehand to further minimize decompression-related delays. The key contributions of this work are summarized as follows:

\begin{itemize}
    \item First, a novel model compression scheme within the hierarchical FL framework is introduced to reduce communication overhead and minimize direct interactions with remote servers.
    \item Second, an adaptive client clustering method selects a core client based on location and residual energy, which then performs local aggregation with compression (LC aggregation).
    \item Third, our model compression approach for hierarchical FL not only cuts communication costs but also enhances decryption and decompression efficiency, making it suitable for broader FL applications without added delays.
\end{itemize}

\par The structure of the paper is organized as follows: Section II discusses related work. Section III outlines the system network. Section IV details the proposed adaptive client clustering algorithm and LC aggregation method. Numerical experiment results are presented in Section V, with concluding remarks in Section VI. 

\section{Related Work}

\subsection{Overview of Federated Learning}
\par The growing scale of data has constrained the development of centralized training paradigms due to substantial data transmission overhead. To counter this, Brisimi \emph{et al.} \cite{brisimi2018federated} introduced a federated optimization scheme for the sparse Support Vector Machine problem, showing that their cluster Primal Dual Splitting algorithm offers improved convergence and reduced communication costs compared to other methods. Similarly, Chen \emph{et al.} \cite{chen2021fedgraph} developed FedGraph for federated graph learning, enhancing privacy protection with a cross-client convolution operation and a sampling algorithm based on deep reinforcement learning to decrease training overhead in Graph Convolutional Networks. While these advancements in FL improve communication efficiency over centralized approaches, the increasing size of neural network models challenges the sustainability of traditional FL mechanisms.

\subsection{Green Communication in Federated Learning}
\par To mitigate the high communication costs in federated learning (FL) training, extensive research has been undertaken. Ghosh \emph{et al.} \cite{ghosh2020efficient} addressed this by clustering clients and introducing an iterative federated clustering algorithm that alternates between identifying user clusters and optimizing model parameters via gradient descent. Hamer \emph{et al.} \cite{hamer2020fedboost} proposed using an ensemble of pre-trained base predictors trained via FL to manage communication bandwidth and storage constraints. Yu \emph{et al.} \cite{yu2022efficient} developed a Multi-Layer FL framework and a corresponding Multi-Layer Stochastic Gradient Descent algorithm to enhance training efficiency and model optimization. While these approaches significantly improve communication efficiency, they focus primarily on reducing transmission frequency rather than model size and require pre-defined cluster numbers. Considering more strategic cluster head selection and cluster design could further decrease energy consumption.

\section{System Model}
\subsection{Hierarchical Federated Learning Framework}
\par In this section, we introduce an efficient model compression strategy for hierarchical FL. As shown in Fig.\ref{system_model}, the system framework includes one server and numerous clients, such as laptops, smart vehicles, and smartphones. In the proposed system, clients store and collect original data locally. Additionally, each cluster within this framework features a core client, details of which will be discussed further.

\par {\textbf{Clients:}} The clients, including user terminals with limited computation and communication capabilities such as smart cars and phones, perform three key functions in our system: 1) They download and train the neural network model using their local data. 2) They send their current position information to the server to assist with clustering and core client selection. 3) They compress the trained model and transmit it to the core client for local aggregation with compression. 

\par Each client communicates with the core client directly, and through the core client sends their local models to the server for aggregation. For ease of illustration, we assume that there are totally ${n_i}$ clients in cluster $i$, $i \in \{1,2,...,t\}$. We represent the data in the \emph{j}-th client in cluster \emph{i}, $j \in \{1,2,...,c\}$ as ${\textbf{D}}_{ij}$ as follows:

\begin{equation}\label{eq:dataset}
    {{\textbf{D}}_{ij}}=\{{{\textbf{X}}_{ij}},{{\textbf{Y}}_{ij}}\},
\end{equation}
where ${\textbf{X}}_{ij}$ represents the data feature set and the corresponding label set is denoted as ${\textbf{Y}}_{ij}$.
\par The \emph{ij}-th client utilizes the local data to train the local model ${\textbf{M}}_{ij}$. First, each model has layers of neural network which can be represented by ${{\textbf{L}}_{ijk}}, k \in \{1,2,...,q\}$ which denotes the \emph{k}-th layer in neural network model of the \emph{j}-th client which belongs to the \emph{i}-th cluster. Besides, each layer ${\textbf{L}}_{ijk}$ contains its parameters and its shape can be arranged as a tensor or a matrix. In other words, the layer ${\textbf{L}}_{ijk}$ consists of neural network parameters that are arranged as follows: 
\begin{equation}
    P_{ijkg},g \in \{1,2,...,G_{k}\},
\end{equation}
where $G_{k}$ represents the number of neural network parameter in the layer ${\textbf{L}}_{ijk}$.
\par After training the neural network with their local datasets, clients compress the model using the proposed method to enhance communication efficiency and privacy. The compressed and perturbed model ${\textbf{M}}'_{ij}$ is then sent to the core client within the same cluster for further processing. Details of this operation are provided in Section IV.

\par {\textbf{Core clients:}} Each cluster selects only one core client, which, in addition to standard client functions, gathers the compressed models from other clients in the same cluster and performs local aggregation with compression (LC aggregation). After that, the core client sends the local aggregated model to the server for further operation. 

\par {\textbf{Server:}} Similar to the conventional FL framework, the server plays four key roles: 1) Adaptive Client Clustering: The server clusters clients based on received positional data. 2) Decompression and Decryption: It performs decompression and decryption on the compressed and aggregated models collected from core clients. 3) Global Aggregation: The server aggregates all the decompressed models using algorithms like FedAvg. 4) Broadcasting: After global aggregation, the server broadcasts the updated global model and compression matrix to all core clients for local model updates and compression. 

\par The main functions of the server will be presented explicitly in Section IV.

\subsection{Problem Definition}
\par The process of hierarchical FL is introduced as shown in Fig.\ref{process}. Assume that there are \emph{T} clients and these clients can be divided into \emph{t} clusters, with $n_i$ clients in each cluster $i$, and each cluster has one core client. 

\par The objective of FL is to train a local model by utilizing the local dataset of clients. The loss function for each client is $l\left( {{{\bf{M}}_{ij}};\left( {{{\bf{X}}_{ij}},{{\bf{Y}}_{ij}}} \right)} \right)$, and the local training is to minimize it as follows:
\begin{equation}\label{clientloss}
    \mathop {\min }\limits_{{{\bf{M}}_{ij}}} l\left( {{{\bf{M}}_{ij}};\left( {{{\bf{X}}_{ij}},{{\bf{Y}}_{ij}}} \right)} \right),
\end{equation}
where ${\textbf{M}}_{ij}$ denotes the model of the \emph{j}-th client in the \emph{i}-th cluster, and the cross-entropy loss function is applied in this work. The most common solution to the problem (\ref{clientloss}) can be achieved by an iterative training process such as gradient descent (GD), stochastic gradient descent (SGD), and so on.
\begin{equation}
    {{\textbf{M}}_{ij}}\left( {\tau  + 1} \right) = {{\textbf{M}}_{ij}}\left( \tau  \right) - r\frac{{\begin{array}{*{20}{c}}
   \partial  & {l\left( {{{\textbf{M}}_{ij}};\left( {{{\textbf{X}}_{ij}},{{\textbf{Y}}_{ij}}} \right)} \right)}  \\
\end{array}}}{{\begin{array}{*{20}{c}}
   \partial  & {{{\textbf{M}}_{ij}}\left( \tau  \right)}  \\
\end{array}}},
\end{equation}
where $\tau$ means the training epoch and $r$ represents the learning rate. 


\par For green communication, we design a local aggregation with a compression method, which avoids clients transmitting their model to the server directly. In each cluster, a core client is selected to receive other clients' models in the same cluster. Then it performs the proposed LC aggregation before global aggregation by the following function.
\begin{equation}
    {\textbf{M}}''_{i}=\sum\limits_{j=1}^{c}  {\textbf{M}}'_{ij},
\end{equation}
where ${\textbf{M}}''_{i}$ is the local compressed aggregated model of the \emph{i}-th cluster. Then, the core client sends ${\textbf{M}}''_{i}$ to the server for further operations.

\par After the server receives all the models from clusters. It performs global aggregation by the following equation.
\begin{equation}\label{eq:globalmodel}
    {{\bf{M}}_{global}} = \sum\limits_{i = 1}^t {{a_i}{{\hat{\textbf{M}}}_i}} ,
\end{equation}
where ${\textbf{M}}_{global}$ is the global aggregated neural network model, and $a_i$ represents the weight of the size of dataset of the \emph{i}-th cluster (e.g., ${a_i} = \left| {{{\textbf{D}}_i}} \right|/\left| {\textbf{D}} \right|$), where ${\textbf{D}}_i$ is dataset of the \emph{i}-th cluster. The ${{\hat{\textbf{M}}}_i}$ represents the decompressed model of the \emph{i}-th cluster.
\par Therefore, the overall objective of FL can be transformed as:
\begin{eqnarray}
 \mathop {\min }\limits_{{{\textbf{M}}_{i}}} \sum\limits_{i = 1}^t { {{a_{i}}l\left( {{{\bf{M}}_{i}};\left( {{{\bf{X}}_{i}},{{\bf{Y}}_{i}}} \right)} \right)}} \notag \\
\text{subject to:} (\ref{eq:dataset}),  (\ref{clientloss})-(\ref{eq:globalmodel}). \notag
\end{eqnarray}

\section{Efficient Model Compression based Hierarchical Federated Learning}
\par
\begin{figure*}[htb]
\centering
\includegraphics[width=5.2in]{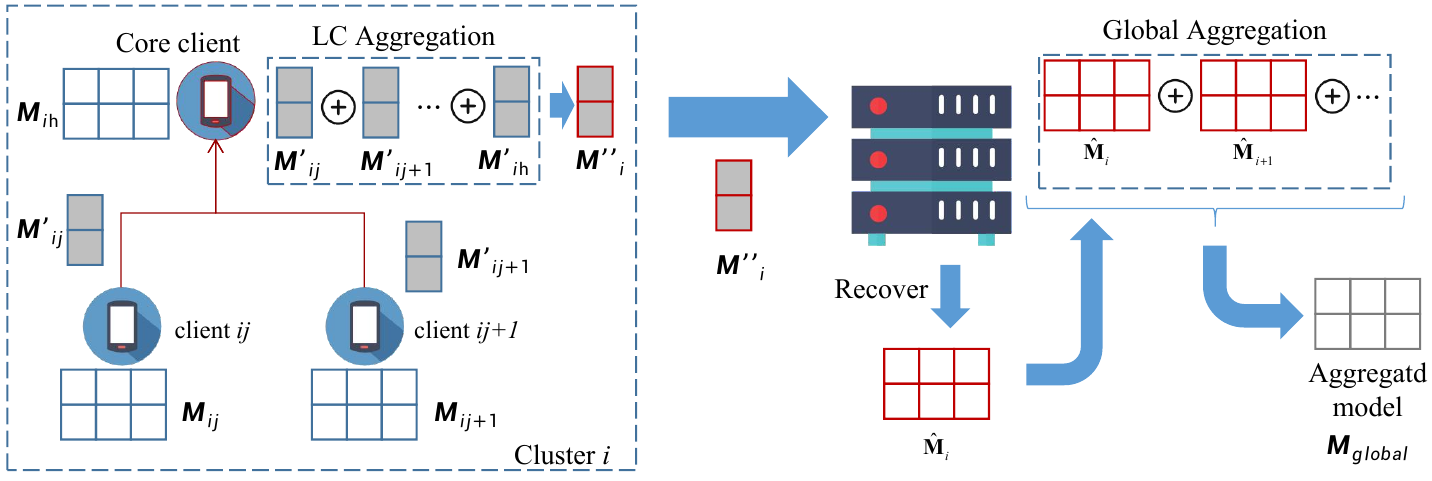}
\caption{The Architecture of Efficient Model Compression based Hierarchical Federated Learning Framework}
\label{process}
\end{figure*}

 In this section, we outline an adaptive client clustering and an efficient model compression \& decompression framework for hierarchical FL. The entire process of our proposed framework is depicted in Fig.\ref{process} and consists of five main stages: client clustering, local training \& compression, LC aggregation, model decompression \& global aggregation, and model broadcasting. 

\subsection{Adaptive Client Clustering}
\par To reduce communication costs and enhance client energy utilization in each cluster, we propose an adaptive client clustering algorithm. Triggered after clients send their location data to the server, the algorithm consists of two main phases: clustering and core client selection. This approach is particularly suited for scenarios where clients, like smartphones and laptops, are spaced several hundred meters apart. 

\par 1) Clustering phase: The server creates 2-D coordinates from the position information uploaded by clients and performs clustering. While methods like K-means are efficient, they require pre-specifying the number of clusters, which can impact clustering effectiveness. Therefore, we employ a DBSCAN-based method to cluster clients based on their geographic distribution. Then, clients are divided into clusters based on their density distribution without pre-setting the number of clusters. However, this method does not automatically identify a suitable core client for each cluster, which could significantly reduce overall cluster costs. To address this, we propose a core client-selecting algorithm.

\par 2) Core client selection phase: After clustering, the server generates a random number $s_{ij}$ (ranging from 0 to 1) and a threshold ${T_{ij}}$ for each client using the following function. 

\begin{equation}\label{clusterhead}
    {T_{ij}} = \frac{p}{{1 - p\left( {u\bmod \frac{1}{p}} \right)}}\frac{{{E_{ij\_rest}}}}{{{E_{avg}}}}\left( {1 - \frac{{{d_{ij\_cen}}}}{{{d_{avg}}}}} \right),
\end{equation}
where $u$ is the current round and $p$ denotes the expected core client percentage. $E_{ij\_rest}$ and $d_{ij\_cen}$ are the \emph{ij}-th client's rest energy and distance between the \emph{ij}-th client and centroid of cluster \emph{i} respectively. The average residual energy ${E_{avg}}$ and average centroid distance ${d_{avg}}$ are defined as follow,

\begin{equation}
    {E_{avg}} = \frac{1}{c}\sum\limits_{j = 1}^c {{E_{ij\_rest}}} ,
\end{equation}
\begin{equation}
    {d_{avg}} = \frac{1}{c}\sum\limits_{j = 1}^c {{d_{ij\_cen}}} .
\end{equation}

\par The function (\ref{clusterhead}) can be divided into three segments. ${p}/{{1 - p\left( {u\bmod \frac{1}{p}} \right)}}$ means randomization. ${{{E_{ij\_rest}}}}/{{{E_{avg}}}}$ represents the level of client's residual energy in cluster. ${{{d_{ij\_rest}}}}/{{{d_{avg}}}}$ describes degree of difference between client location and cluster center. Therefore, our method takes the residual energy of the clients in the cluster and their location into account.

\par If $s_{ij}<T_{ij}$, the \emph{ij}-th client will be selected as a core client. Conversely, the \emph{ij}-th client will be not selected as a core client. By utilizing this adaptive client clustering algorithm, the client that is closer to the centroid and has more residual energy is more likely to be selected as the core client.

\subsection{Model Compression and LC aggregation}
\par Model compression involves two steps: model sparsification (or pruning) and perturbed model compression, both reducing size while protecting privacy. The neural network model is compressed layer by layer, except for the output layer, which contains more personal information and enhances local dataset performance \cite{collins2021exploiting}. Thus, the output layer remains with the clients locally. 

\par 1) Model Sparsification: Empirical experiments \cite{wu2022improved} show that many neural network parameters are redundant and unimportant for overall performance. However, these unnecessary parameters significantly increase communication consumption. 
\par Therefore, a threshold $\mu$ can be set to filter these unimportant parameters, which is defined as:
\begin{equation}
    {P^{sparse}_{ijkg}} = \left\{ {\begin{array}{*{20}{c}}
   {{P_{ijkg}},} & {\left| {{P_{ijkg}}} \right| \ge \mu }  \\
   {0,} & {\left| {{P_{ijkg}}} \right| < \mu }  \\
\end{array}} \right..
\end{equation}

\par After performing model sparsification, the raw layer ${\textbf{L}}_{ijk}$ will become sparse layer ${\textbf{L}}^{sparse}_{ijk}, k\in\{1,2,...,q\}$.

\par 2) Perturbed Model Compression: While model sparsification reduces model size by pruning unimportant parameters, it does not address privacy preservation and explore coherence among parameters. Therefore, the perturb model compression is utilized to solve these two problems mentioned above. First, each sparse layer of model is reshaped from tensor or matrix to vector i.e., reshaping ${\textbf{L}}^{sparse}_{ijk}$ to ${\tilde{\textbf{L}}}_{ijk}^{sparse} \in {R^{{G_k} \times 1}}$. Then, each client utilizes a compression matrix to compress the model layer by layer via the following equation, which is distributed by the server after finishing the clustering. In addition, the same compression matrix is used for the same cluster. 
\par 
\begin{equation}\label{comp}
    {{\textbf{L}}'_{ijk}} = {{\bf{\Phi }}_{ik}}\cdot{\bf{\tilde L}}_{ijk}^{sparse},
\end{equation}
where ${{\bf{\Phi }}_{ik}} \in {R^{{g_k} \times G_k}},({g_k} \ll {G_k})$ is the compression matrix of the \emph{k}-th layer of any clients in the \emph{i}-th cluster. Thus, the compressed model is changed as follows:

\begin{equation}
    {{\textbf{M}}'_{ij}} = \left\{ {{{{\bf{L'}}}_{ij1}},{{{\bf{L'}}}_{ij2}}, \ldots ,{{{\bf{L'}}}_{ijq}}} \right\}.
\end{equation}

\par After performing these two model compression, clients will send the compressed model ${{\textbf{M}}'_{ij}}$ to the corresponding core client of the cluster for LC aggregation. 

\par When LC aggregation complete, the compressed model ${\textbf{M}}''_{i}$ consists of many aggregated layers, which are arranged as:
\begin{equation}
    {{\bf{L''}}_{ik}} = \sum\limits_{j = 1}^c {{{{\bf{L'}}}_{ijk}}} ,k \in \{ 1,2, \ldots ,q\} ,
\end{equation}

\begin{equation}
    {{\textbf{M}}''_{i}} = \left\{ {{{{\bf{L''}}}_{i1}},{{{\bf{L''}}}_{i2}}, \ldots ,{{{\bf{L''}}}_{iq}}} \right\}.
\end{equation}

\subsection{Efficient Model Decompression and Decryption}


\par In our previous work \cite{zhu2023model}, each client connects to the server directly. So the server can only decompress clients' compressed models one by one via solving the following optimization problem {\bf{several times}} which depends on the number of clients. 

\begin{equation}
    \begin{array}{*{20}{c}}
   {{\bf{P1}}:} & {\begin{array}{*{20}{c}}
   {\mathop {\min }\limits_{{\bf{\tilde L}}_{ijk}^{sparse}} } & {{{\left\| {{\bf{\tilde L}}_{ijk}^{sparse}} \right\|}_1}}  \\
\end{array}}  \\
   {s.t.} & {{{{\bf{L''}}}_{ijk}} = {{\bf{\Phi }}_{ik}} \cdot {\bf{\tilde L}}_{ijk}^{sparse}}  \\
\end{array}.
\end{equation}

\par However, in this paper, an efficient and equivalent decompression is proposed to decompress all compressed models cluster by cluster.
\par When the server receives the required data (i.e., LC aggregated neural network models ${\textbf{M}}''_{i}$) from all clusters, it will solve the underdetermined system of equations (\ref{comp}) by the following optimization problem.

\begin{equation}
    \begin{array}{*{20}{c}}
   {{\bf{P2}}:} & {\mathop {\min }\limits_{{\bf{\tilde L}}^{sparse}_{ik}} {{\left\| {{\bf{\tilde L}}^{sparse}_{ik}} \right\|}_1}}  \\
   {s.t.} & {{{\bf{L}}''}_{ik} = {{\bf{\Phi }}_{ik}}\cdot{\bf{\tilde L}}^{sparse}_{ik}}  \\
\end{array},
\end{equation}
where ${\bf{\tilde L}}^{sparse}_{ik} = \sum\limits_{j = 1}^c {{\bf{\tilde L}}_{ijk}^{sparse}}$ and ${\bf{L}}''_{ik} = \sum\limits_{j = 1}^c {{\bf{L}}''_{ijk}}$ denote the aggregated \emph{k}-th layer which is pruned and perturbed of all clients in the \emph{i}-th cluster respectively.

\par {\bf{Theorem 1:}} \emph{Solving the optimization problem {\textbf{P1}} multiple times is equivalent to solving {\textbf{P2}} once.}

\par {\bf{Proof:}} For convenience, we assume that there are two clients, and their pruned models are defined as {\bf{A}} and {\bf{B}}. The compression matrix is denoted as {\bf{C}}. 

\par After performing the perturbed model compression method, their model will be changed as follow.
\begin{equation}
    {\textbf{A}}'={\textbf{C}}{\textbf{A}},
    {\textbf{B}}'={\textbf{C}}{\textbf{B}},
\end{equation}

\par Then LC aggregation is executed:
\begin{equation}\label{yasuo}
    \begin{array}{c}
 {{\bf{A}}^\prime } + {{\bf{B}}^\prime } = {\bf{CA}} + {\bf{CB}} \\ 
  = {\bf{C}}\left( {{\bf{A}} + {\bf{B}}} \right) \\ 
 \end{array}
\end{equation}

\par The ultimate goal of federal learning is obtaining the aggregated model (i.e., ${\bf{A}} + {\bf{B}}$). So, we let ${\bf{D}}={\bf{A}} + {\bf{B}}$ and ${\bf{D}}'={\bf{A}}' + {\bf{B}}'$, and the equation (\ref{yasuo}) can be transformed as:
\begin{equation}
    {\textbf{D}}'={\textbf{C}}{\textbf{D}}.
\end{equation}
Therefore, the target problem is solving ${\textbf{D}}$ when ${\textbf{C}}$ and ${\textbf{D}}'$ are known, which can be solved by the following optimization.
\begin{equation}
    \begin{array}{*{20}{c}}
   {} & {\begin{array}{*{20}{c}}
   {\mathop {\min }\limits_{\bf{D}} } & {{{\left\| {\bf{D}} \right\|}_1}}  \\
\end{array}}  \\
   {s.t.} & {{\bf{D'}} = {\bf{C}} \cdot {\bf{D}}} \\
\end{array}.
\end{equation}
Therefore, when ${\bf{A}}' + {\bf{B}}'$ and $\textbf{C}$ are known, ${\bf{A}} + {\bf{B}}$ can be obtained without solving ${\bf{A}}$ and ${\bf{B}}$ respectively.

\section{Performance Evaluation}
\subsection{Experimental Setting}
\par We evaluate our method using a CNN model commonly used in object recognition tasks. It features two convolutional layers with ReLU activation and a fully connected output layer with a softmax layer. We tested the hierarchical FL framework on two datasets, MINIST and CIFAR-10, using a system setup of 30 clients and one server.

\par For comparison, we consider two baseline methods: one for model compression and the other based on a clustered FL framework. The first, 'ClusterSpar' \cite{cui2020clustergrad}, uses the K-means algorithm to identify key gradients distant from zero, which are then approximated using a clustering-based quantization technique. The second, 'ML-FL' \cite{yu2022efficient}, involves a multi-layer FL framework that aggregates partial models at edge nodes before exchanging intermediate results with the server, facilitating efficient and rapid global model training. 

\par For all experiments, the learning rate is set as 0.01. In order to be close to the actual scenario, the radius of the neighborhood is set as $r\_neighbor = 5$km, and the threshold of the number of clients in the neighborhood $p\_min$ is 5. Therefore, the energy cost can be calculated as follows 
\begin{equation}\label{ecost}
    \begin{array}{c}
 {{\mathop{\rm cost}\nolimits} _{total}} = {{\mathop{\rm cost}\nolimits} _{client}} + {{\mathop{\rm cost}\nolimits} _{core}} \\ 
  = {p_{cluster}}\cdot{t_{cluster}} + {p_{server}}\cdot{t_{server}} \\ 
  = {p_{cluster}}\cdot\frac{{{S_{ij}}}}{v_{cluster}} + {p_{server}}\cdot\frac{{{S_i}}}{v_{server}}, \\
 \end{array}
\end{equation}
where ${{\mathop{\rm cost}\nolimits} _{client}}$ and ${{\mathop{\rm cost}\nolimits} _{core}}$ represent the energy cost of common client in a cluster and core client respectively. ${p_{cluster}}$ is the transmission power between the client and core client in a cluster, while ${p_{server}}$ denotes the transmission power between server and core client which are set as a constant number $10^{-4}$W and $10^{-2}$W respectively. ${v_{cluster}}$ and ${v_{server}}$ mean the data transmission speed among the cluster and the data transmission speed between the core client \& server. $S_{i}$ and $S_{ij}$ represent the size of model of the $i$-th cluster and the $j$-th client in the $i$-th cluster respectively.
\par In addition, ${v_{cluster}}$ and ${v_{server}}$ are defined as:
\begin{equation}
    {v_{cluster}} = B\cdot{\log _2}(1 + \frac{{{h^2}{p_{cluster}}}}{{{N_0}}}),
\end{equation}
\begin{equation}
    {v_{server}} = B\cdot{\log _2}(1 + \frac{{{h^2}{p_{server}}}}{{{N_0}}}),
\end{equation}
where $B = 40$Mb/s is bandwidth and $N_0 = 10^{-10}$W means the power of noise. $h$ denotes wireless channel gain, which is set to 1.

\subsection{Experimental results}
\begin{figure}[htb]
\begin{subfigure}{0.24 \textwidth}
   \includegraphics[width=1.\linewidth]{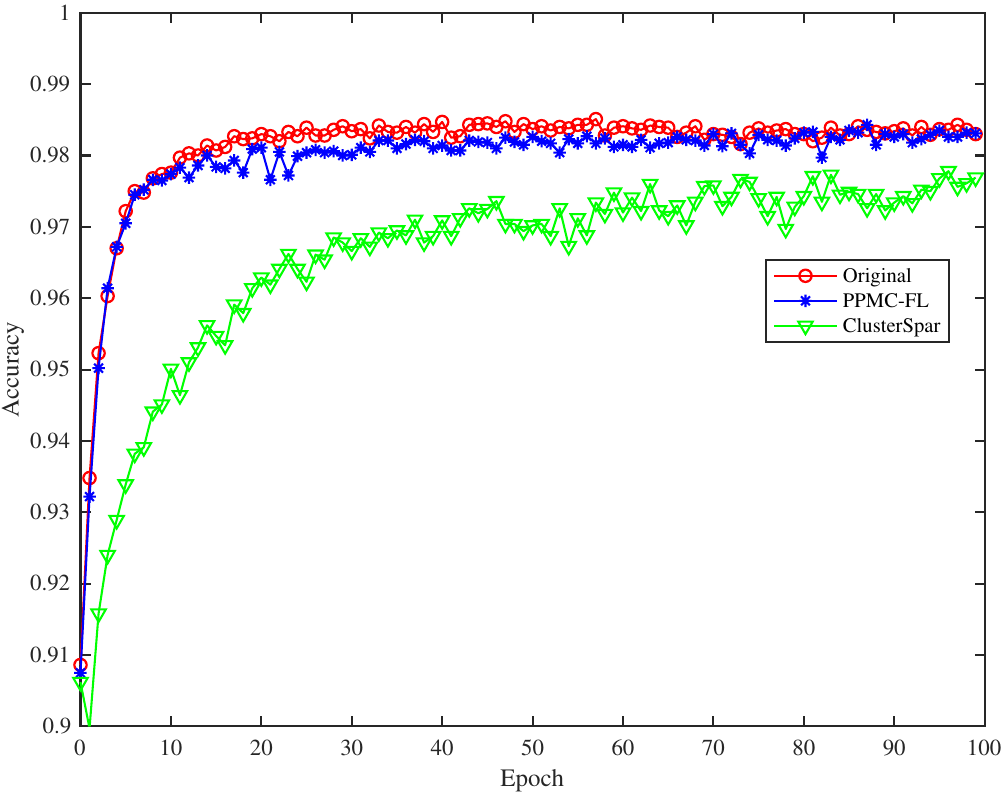} 
\caption{Accuracy on MNIST}
\label{fig:subim1} 
\end{subfigure}
\begin{subfigure}{0.24 \textwidth}
   \includegraphics[width=1.\linewidth]{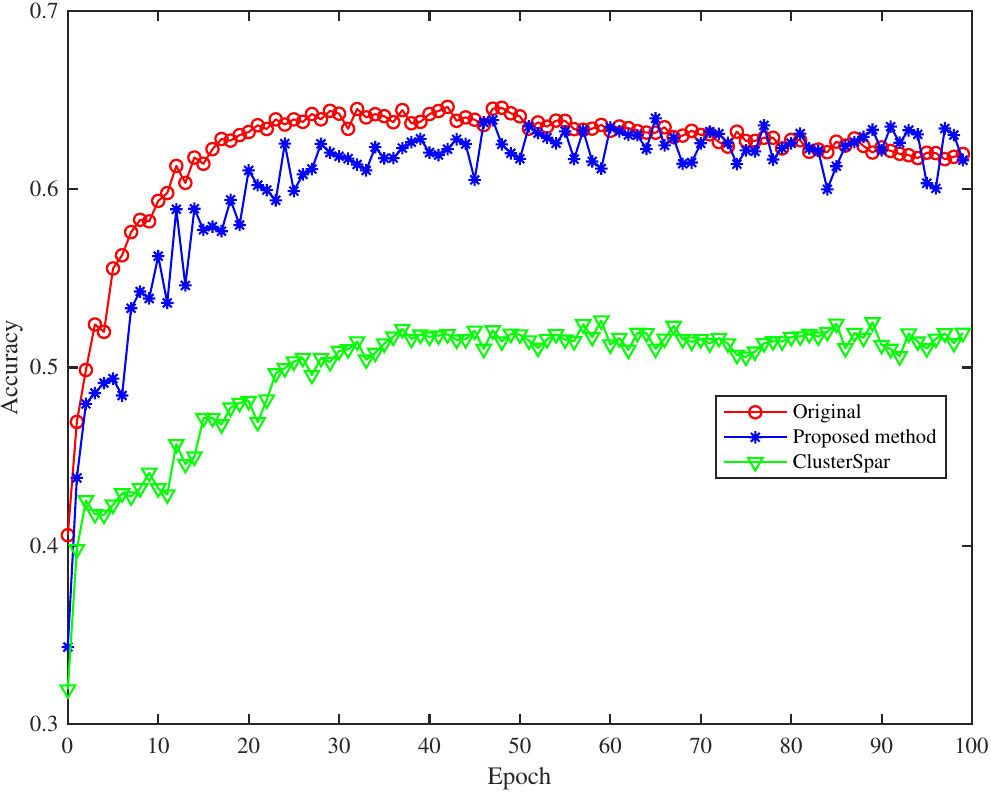} 
\caption{Accuracy on CIFAR}
\label{fig:subim1} 
\end{subfigure}
    \caption{Accuracy with different method on two datasets}
    \label{comratio}
\end{figure}

\par Fig. \ref{comratio} displays the prediction accuracy of various methods on the MNIST and CIFAR-10 datasets across 100 iterative aggregations (including LC and global aggregation). All compression techniques are applied post-local training at the client side. 'Original' refers to no compression use, similar to the 'ML-FL' method, which doesn't utilize model compression and thus doesn't affect prediction accuracy. ClusterSpar' refers to the method proposed in the work \cite{cui2020clustergrad}. 'Proposed method' indicates our method. From the figure, we observe that all methods show increasing accuracy over training epochs, with our proposed method achieving the highest performance on both datasets, demonstrating that it effectively reduces model size while maintaining accuracy. 


\begin{figure}[htb]
\begin{subfigure}{0.24 \textwidth}
   \includegraphics[width=1.\linewidth]{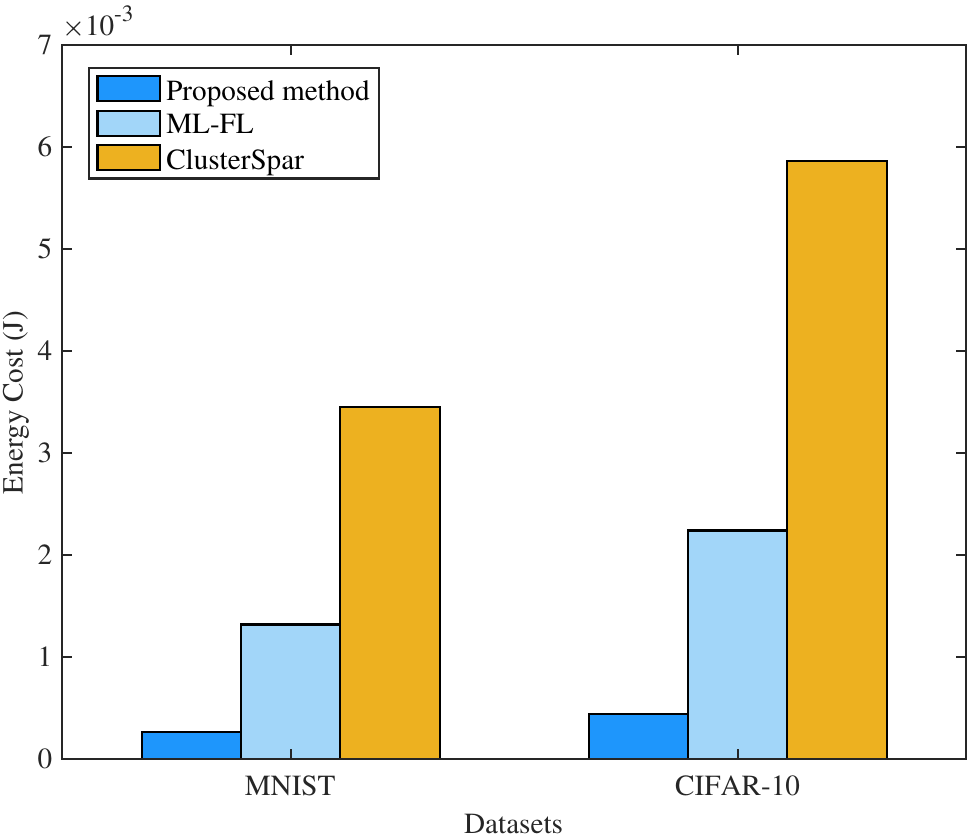} 
\caption{Energy consumption among different scheme on two datasets}
\label{cost} 
\end{subfigure}
\begin{subfigure}{0.24 \textwidth}
   \includegraphics[width=1.\linewidth]{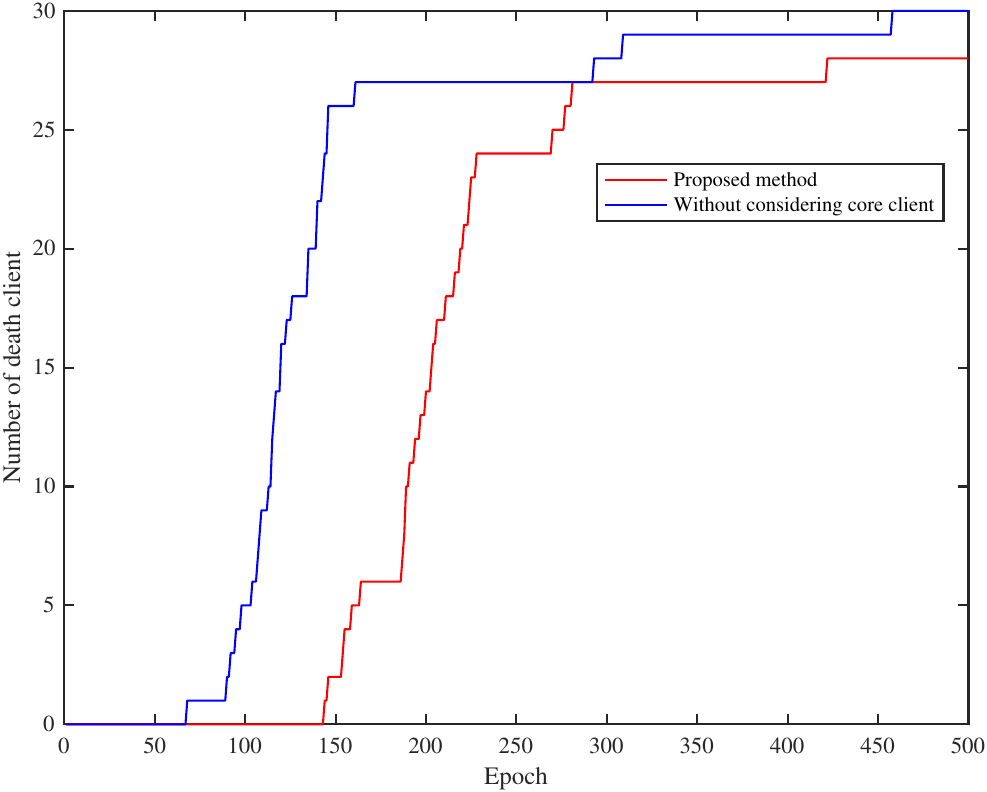} 
\caption{Impact of core-client selection algorithm on energy cost}
\label{dn} 
\end{subfigure}
    \caption{Impact of dataset \& core-client selection on cost}
    \label{total_cost}
\end{figure}

\par Fig. \ref{cost} shows the energy consumption of three different schemes, calculated using equation (\ref{ecost}). The proposed scheme records the lowest energy consumption on both datasets. To understand this cost advantage, we analyze the number of clients with depleted energy after 500 epochs of global aggregation in Fig. \ref{dn}. This analysis reveals that methods neglecting core client selection rapidly deplete certain clients' energy. Conversely, the proposed scheme selects core clients more effectively in each cluster, thereby balancing energy costs among clients. 


\begin{figure}[htb]
\centering
\includegraphics[width=2.5in]{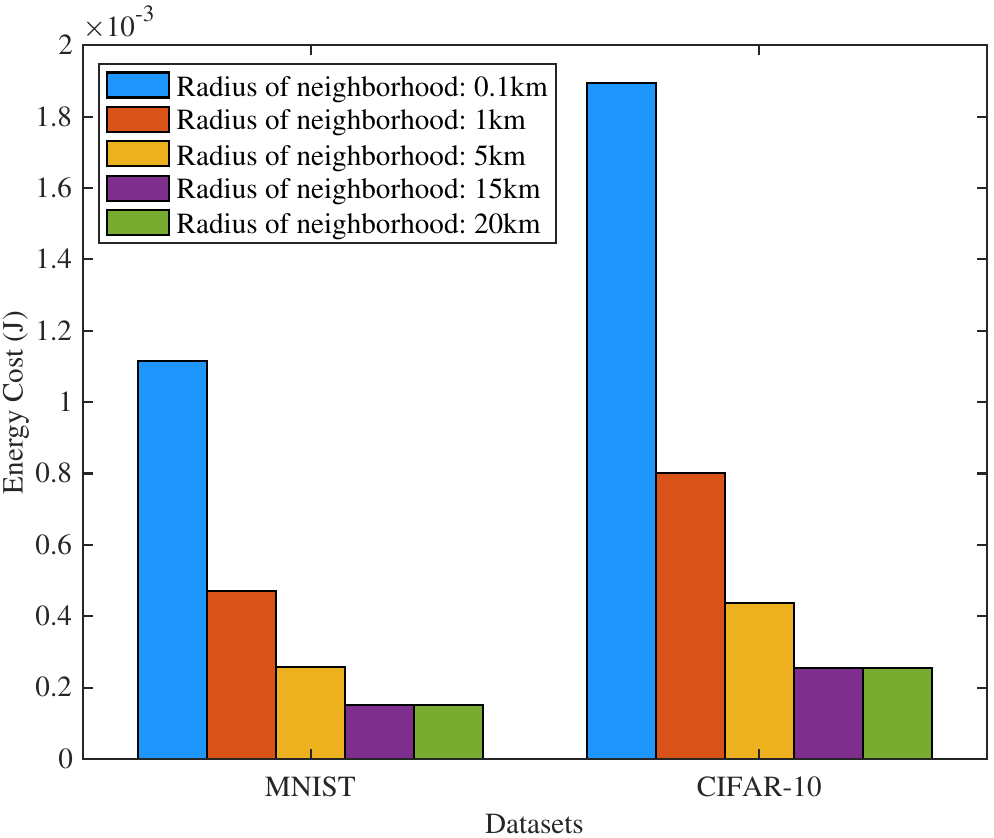}
\caption{Influence of radius of neighborhood on energy cost}
\label{dr}
\end{figure}
\par Fig. \ref{dr} illustrates how the neighborhood radius ($r\_neighbor$) affects energy consumption. The figure shows that energy consumption decreases as the neighborhood radius increases but eventually stabilizes. This stabilization occurs because once the radius expands sufficiently, all clients are grouped into a single cluster. 

\section{Conclusions}
\par In this paper, we introduce a novel model compression scheme within a hierarchical federated learning framework. This framework prevents direct client-server connections, balances energy costs across clusters with local aggregation and compression, and uses an adaptive client clustering method. We also analyze improvements in decompression and decryption theoretically. Our experiments demonstrate that the proposed scheme reduces energy consumption without sacrificing prediction accuracy. Future work will explore optimizing compression ratios across different clusters to balance energy consumption and accuracy. 

\section*{Acknowledgment}
\par This work is supported by National Nature Science Foundation of China, No. 62072485, and Guangdong Basic and Applied Basic Research Foundation No. 2022A1515011294.

\bibliographystyle{unsrt} 
\bibliography{zx.bib}

\end{document}